\documentclass[conference]{IEEEtran}
\IEEEoverridecommandlockouts
\usepackage{cite}
\usepackage{amsmath,amssymb,amsfonts}
\usepackage[noend]{algpseudocode}
\usepackage{algorithmicx,algorithm}
\usepackage{graphicx}
\usepackage{textcomp}
\usepackage{xcolor}
\usepackage{booktabs}
\usepackage{multirow}
\usepackage{float}
\usepackage{bigstrut}

\newcommand{\tabincell}[2]{\begin{tabular}{@{}#1@{}}#2\end{tabular}}

\def\BibTeX{{\rm B\kern-.05em{\sc i\kern-.025em b}\kern-.08em
    T\kern-.1667em\lower.7ex\hbox{E}\kern-.125emX}}

\begin{document}

\title{AWCD: An Efficient Point Cloud Processing Approach via Wasserstein Curvature}

\author{
	 Yihao Luo\textsuperscript{1}, Ailing Yang\textsuperscript{2} , Fupeng Sun\textsuperscript{1}   , Huafei Sun\textsuperscript{$\ast$\ 1}   \\
	{\small \it 1. School of Mathematics and Statistics, Beijing Institute of Technology, Beijing, China}\\
	{\small \it 2. School of Statistics, University of International Business and Economics, Beijing, China }\\
{\small knowthingless@bit.edu.cn\ 201856024@uibe.edu.cn\  fupeng\_sun@bit.edu.cn\ huafeisun@bit.edu.cn}\\
{\small \it Corresponding Author: Huafei Sun    Email:huafeisun@bit.edu.cn}
}

%
%
%
%
%
%

\maketitle

\begin{abstract}
In this paper, we introduce the adaptive Wasserstein curvature denoising (AWCD), an original processing approach for point cloud data. By collecting curvatures information from Wasserstein distance, AWCD consider more precise structures of data and preserves stability and effectiveness even for data with noise in high density. This paper contains some theoretical analysis about the Wasserstein curvature and the complete algorithm of AWCD. In addition, we design digital experiments to show the denoising effect of AWCD. According to comparison results, we present the advantages of AWCD against traditional algorithms.
\end{abstract}

\begin{IEEEkeywords}
point cloud, denoising, Wasserstein distance, curvature
\end{IEEEkeywords}

\section{Introduction}
Along the blooming of data science, point cloud processing, especially denoising, plays an increasingly important role in data relevant researches and engineering. Denoising or outliers removing is always the first and one of the most essential step of the data processing as a result of the common pollution of data from noise. However, the noise in high density is usually hard to be removed.\par
In industry, PCL\cite{ref:PCL} is a developed and widely used platform for point cloud processing. This platform collects four main denoising approaches, 
but neither of them is efficient for data with dense noise. To solve this problem, we tend to consider more precise structure of data and design a new algorithm.
The starting point is to embedding the original point cloud from $n$-dimensional Euclidean space into a more complicated manifold \cite{ref:Manifold}, corresponding all $n$-variate null-expectation Gaussian distribution, called $SPD(n)$ \cite{ref:PositiveDefiniteMatrices}, through local statistical as well. When $SPD(n)$ is endowed with a natural Riemannian metric \cite{ref:WasserMetric} induced by the famous Wasserstein distance \cite{ref:WassersteinDorgin}, we can determine various curvatures \cite{ref:Curvaturetextbook}, called Wasserstein curvatures. The main results of the geometry of Wasserstein metric on $SPD(n)$ are contained in \cite{ref:Mypaper}. A principal idea of this paper is that the Wasserstein curvatures, especially the Wasserstein scalar curvatures \cite{ref:Curvaturetextbook}, imply the informational quantity. Therefore the first two steps of our algorithm are local statistics and the Wasserstein curvature comuputing.\par
On the other hand, almost all traditional denoising methods are dependent on sorts of artificial parameters. There exists the void of a adaptive denoising method independent on redundant parameters. Actually, this problem can be solved by some brief reorganization of the curvature information. Due to some good properties of the Wasserstein curvature, the curvature of different data clusters can be easily distinguished by the wave crest and wave trough from a histogram of the curvature value. In this way, we design the adaptive denoising approach via the Wasserstein curvature. Hence it is called adaptive Wasserstein curvature denoising (AWCD).\par
In addition, we present the denoising effects of AWCD over diverse data sets, with comparison to two classical methods. According to our results from digital experiments, AWCD holds an obvious advantage regardless of the data size and the density of noise. Besides, an elemental algorithm analysis points that the computational complexity of AWCD would not be higher than traditional ways in a large degree.
Although this paper presents AWCD merely as a prototype, the current algorithm can promote point cloud denoising in some sensible senses.\par
This paper is organized as follows. We tend to introduce the Wasserstein geometry of $SPD(n)$ in section 2 and classical point cloud denoising methods in section 3. Our main results and algorithm are presented in section 4. Section 5 is about digital experiments with the effects comparison. Finally, section 6 contains conclusions and a brief discussion.

\section{Wasserstein Geometry on $SPD(n)$}
Wasserstein distance  \cite{ref:WassersteinDorgin}, called Earth-moving distance as well, describes the minimal price or energy required to move one distribution to another along an underlying manifold. For any two distributions $P_1,P_2$ on manifold $M$, $\Pi\left(P_1,P_2\right)$ denotes the class of all union distributions with $P_1,P_2$ as marginal distributions. Then we have Wasserstein distance from $P_1$ to $P_2$,
\begin{equation}\label{Wsst}
	W_p\left(P_1,P_2\right)= \inf_{\gamma\sim\Pi\left(P_1,P_2\right)} \left(E_{\left(x,y\right)\sim\gamma}[\|x-y\|^p] \right)^{\frac{1}{p}},
\end{equation}
where $E$ is the expectation. The general Wasserstein distance can not be expressed explicitly, unless on some specific distribution families.
Fortunately, the distance among Gaussian distributions on $\mathbb{R}^n$ has a beautiful explicit form  \cite{ref:WasserGGGG}
\begin{equation}\label{Wss2d} W\left(\mathcal{N}_1,\mathcal{N}_2\right)=\|\mu_1-\mu_2\|+ {\rm tr}^\frac{1}{2}(\Sigma_1+\Sigma_2-2\left(\Sigma_1\Sigma_2\right)^{\frac{1}{2}}) ,
\end{equation}
where $\mu_1$, $\mu_2$ and $\Sigma_1$, $\Sigma_1$ are expectations and covariances of Gaussian distributions $\mathcal{N}_1$, $\mathcal{N}_2$ respectively.\par
Wasserstein distance can be induced by a Riemannian metric defined on manifold $SPD(n)$ of $n$-dimensional symmetric positive definite matrixes.	For any ${A}\in SPD\left(n\right)$, and tangent vectors $X,Y\in T_{A}SPD\left(n\right)$,
\begin{equation}\label{Gw metric}
		g_W|_{A}\left(X,Y\right) = \frac{1}{2}{\rm tr}\left(\Gamma_{A}[Y]  X\right),
\end{equation}
where $\Gamma_{A}[Y]:=T$ denotes the solution of Sylvester equation \cite{ref:Sylvester}
\begin{equation}\label{Sylv}
	{A}T+T{A}={Y}.
\end{equation}
We call $g_W$  Wasserstein metric because the geodesic distance of metric \eqref{Gw metric} coincides with Wasserstein distance \eqref{Wss2d}. We denote the Riemannian manifold as $(SPD(n),g_W)$.\par
The authors of \cite{ref:Mypaper} studied various geometric characteristics of $(SPD(n),g_W)$, including geodesics, exponential maps, the Riemannian connection, Jacobi fields and curvatures. This paper mainly concentrates on the curvature information. In the following, we shall present some relevant results.\par
	For any ${A} \in (SPD(n),g_W)$, and ${X},{Y}$ are smooth vector fields on $SPD\left(n\right)$ , Riemannian curvature tensor $\langle R_{{X}{Y}}{X},{Y}\rangle$ at ${A}$ has an explicit expression
\begin{equation}\label{Riecurvature}
	\begin{split}
	    R\left({X},{Y},{X},{Y}\right)
	  =3{\rm tr}  & (\Gamma_{A}[{X}]{A}\Gamma_{A}[\Gamma_{A}[{X}]\Gamma_{A}[{Y}]\\
&-\Gamma_{A}[{Y}]\Gamma_{A}[{X}]]{A}\Gamma_{A}[{Y}]).
	\end{split}
\end{equation}
Subsequently, we have the explicit expression of scalar curvature.
For any ${A} \in SPD\left(n\right)$, the scalar curvature satisfies
\begin{equation}\label{Curve scalar}
\begin{split}
   &\rho\left({A}\right) := \sum_{i=1}^{n}\sum_{j=1}^{n}R\left({e_i},{e_j},{e_i},{e_j}\right)\\
    = & 3{\rm tr}\left(2\Lambda U U^{T}+\Lambda U^{T}U+\Lambda U\left(U+U^{T}\right)\Lambda \left(U+U^{T}\right)\right),
\end{split}
\end{equation}
	where $\{e_i\}$ is any orthonormal basis of $T_A{SPD(n)}$   , $\Lambda= {\rm diag}[\lambda_{1}, \cdot\cdot\cdot, \lambda_{n}]$ as a diagonalization of ${A}$, and $U:=[\frac{1}{\lambda_{i}+\lambda_{j}}]_{i<j}$ is a upper triangle matrix. The subscripts of eigenvalue $\lambda$ must keep consistent when constructing $\Lambda$ and $U$. \par The scalar curvature is a point-wise defined function, which will be used essentially in AWCD algorithm. \par
Some properties of the Wasserstein curvature will take benefits for our algorithm as well. To estimate the wasserstein curvature, we have that there exists an orthonormal basis $\{e_k\}$ for $T_{A}SPD\left(n\right)$ such that for any $e_{k_1},e_{k_2} \in \{e_k\}$
\begin{equation} 0<K_{A}\left(e_{k_1},e_{k_2}\right):=\sum_{j=1}^{n}R\left({e_{k_1}},{e_{k_j}},{e_{k_2}},{e_{k_j}}\right)< \frac{3}{\lambda_{min2}\left({A}\right)},
\end{equation}
where $\lambda_{min2}$ is the secondly minimal eigenvalue. From this inequality, we gain
\begin{equation}\label{Scinequal}
 0<\rho\left({A}\right)< \frac{3n}{\lambda_{min2}\left({A}\right)},
\end{equation}
which says that the Wasserstein scalar curvatures are controlled by the secondly minimal eigenvalue. Therefore the curvature will seldom explode even on a domain almost degenerated. Only when the matrices degenerate at over two dimensions will the curvatures be very large. This phoneme ensures the curvature information make sense in most applications. Such an idea is actually the start point of our algorithm.

\section{Traditional Denoising Methods}
\begin{figure}[ht]
\centerline{\includegraphics[width=0.8\linewidth]{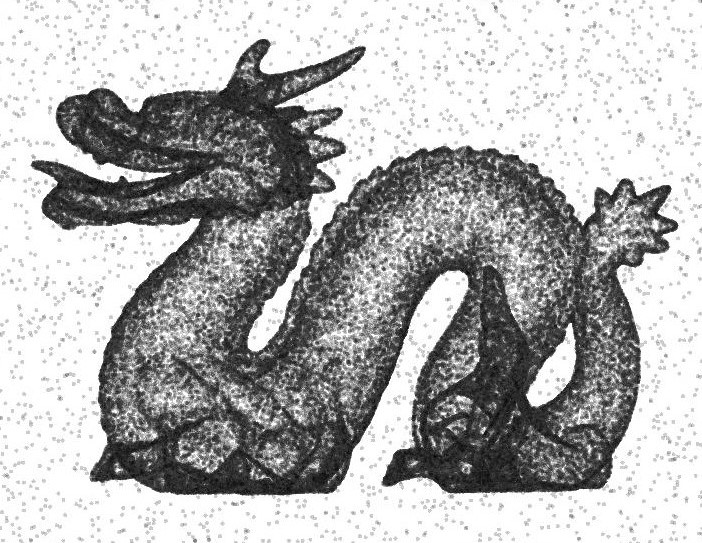}}
\caption{Polluted Point Cloud as Input.}
\label{fig:Origin}
\end{figure}

In this and next sections, we attempt to introduce some algorithms. For convenience, we involve a polluted point cloud (dragon with size of 50000 points) as the example. We will take this point cloud as input for following algorithms. See Fig.~\ref{fig:Origin}.  \par
\par
We will show the effects of following three algorithms in Fig.\ref{fig:ROR}, Fig.\ref{fig:SOR}, and Fig.\ref{fig:AWCD} respectively. We shall remind that the real outputs of those algorithms are unclassified point clouds without coloring. We present the results in colored figures so as to enhance the intuitive feeling for readers. Therefore, we use blue points for real true information, red points to mark false positive noises and yellow points (hard to see perhaps) for false negative real information. \par
PCL \cite{ref:PCL} provides four main point cloud denoising methods. Conditional removal needs to settle artificial conditions and Passthough is just a linear intercept. Due to the lack of their universal usage, we will not focus on above two methods but  radius outlier removal (ROR) and statistical outlier removal (SOR).
\subsection{Radius Outlier Removal}
The idea of ROR is quit elemental, which believes local points density indicating the informational quantity, i.e., a point whose neighbour with less points tends to be recognized as a noise. Therefore, ROR contains two steps of $d$-radius  neighbors searching and counting; There are amounts of developed algorithms for $d$-radius neighbors searching, including a fast approach of KD-Tree \cite{ref:KTTree}. ROR requires two parameters, including $d$ as the radius for local neighborhoods and $\alpha$ as the least threshold of points number in a neighborhood.

\begin{algorithm}[ht]
\label{alg:ROR}
\caption{radius outlier removal} 
\hspace*{0.02in} {\bf Input:} 
initial point cloud $D_0$; parameters $d$, $\alpha$\\
\hspace*{0.02in} {\bf Output:} cleaned cloud $D_1$
\begin{algorithmic}[1]
\State search R-radius neighborhood $N_i$ for each point $P_i$, where $N_i=\{P_j|\|P_i-P_j\|\leq d\}$
\If {number of neighbors $|N_j| \geq \alpha$}{\ put $P_i$ into $D_1$}
\EndIf
\State \Return $D_1$
\end{algorithmic}
\end{algorithm}
\par
\begin{figure}[ht]
\centerline{\includegraphics[width=0.95\linewidth]{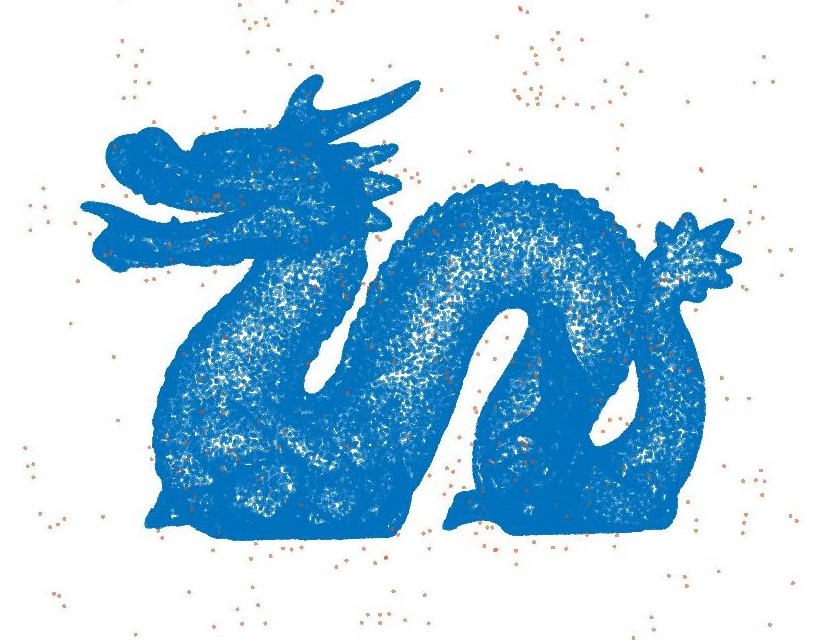}}
\caption{Cleaned Point Cloud as Output of ROR.}
\label{fig:ROR}
\end{figure}

In our example, the effect of ROR is presented in Fig.~\ref{fig:ROR}. It is easily to observe that ROR preserves almost all real information well but remains a considerable number of noises as well.\par
Actually, ROR depends on parameters chosen sensitively. However, there exists no universal method to determine the best parameters. And the complexity of ROR approximates to the complexity of  $O(\log^{2}n+n)$, where $O(\log^{2}n)$ for neighbors searching by KT-Tree and $O(n)$ for point-wise judgements. For details, see \cite{ref:NN}.

\subsection{Statistical Outlier Removal}
Compared to ROR, Statistical Outlier Removal (SOR) is a more precise algorithms which considers the locally statistical structure rather than counting. SOR is one of the most popular methods to pre-treat point clouds, due to its efficiency over data with noises in low density. On the other hand, SOR is almost adaptive with a canonical observation from classical statistics, called one-sigma law \cite{ref:Multivariate}. However, SOR is sensitive to the density of noises. SOR is so  weak to handle dense noises while ROR runs further better.\par
SOR believes a outlier is far from the center of its k-nearest neighborhood. Conversely, a point from the real information always lie in a confidence area of its neighborhood. For a point $P \in \mathbb{R}^n$ and a multivariate Gaussian distribution $\Phi$ with expectation $\mu$ and covariance $\Sigma$, we say $P$ lying the confidence area of $\Phi$  if
\begin{equation}\label{multcofidence}
  {(P-\mu)}^T{\Sigma}{(P-\mu)}\geq{\|P-\mu\|}^4.
\end{equation}
This inequality is a generalization of one-sigma law in higher dimension.\par
By this way, SOR contains three main steps. The fist step is to search k-nearest neighbours (kNN) for every point. The second step is to determine the local Gaussian distribution $\Phi$ for every point, according to the mean and covariance of neighbors. Finally, judge outliers as noises with  \eqref{multcofidence}. SOR requires a single parameter neighbours number $k$ for kNN.
\begin{algorithm}[ht]
\label{alg:SOR}
\caption{statistical outlier removal} 
\hspace*{0.02in} {\bf Input:} 
initial point cloud $D_0$; parameter $k$\\
\hspace*{0.02in} {\bf Output:} cleaned cloud $D_1$
\begin{algorithmic}[1]
\State search kNN $N_i$ for each point $P_i$
\State compute local mean $\mu_i = \frac{1}{k}\sum_{j=1}^{k}N_{ij}$ and local covariance $\Sigma_i = (N_i)^TN_i$
\If {${(P_i-\mu_i)}^T{\Sigma}{(P_i-\mu_i)}\geq{\|P_i-\mu_i\|}^4$}{ put $P_i$ into $D_1$}
\EndIf
\State \Return $D_1$
\end{algorithmic}
\end{algorithm}
\par
\begin{figure}[ht]
\centerline{\includegraphics[width=0.8\linewidth]{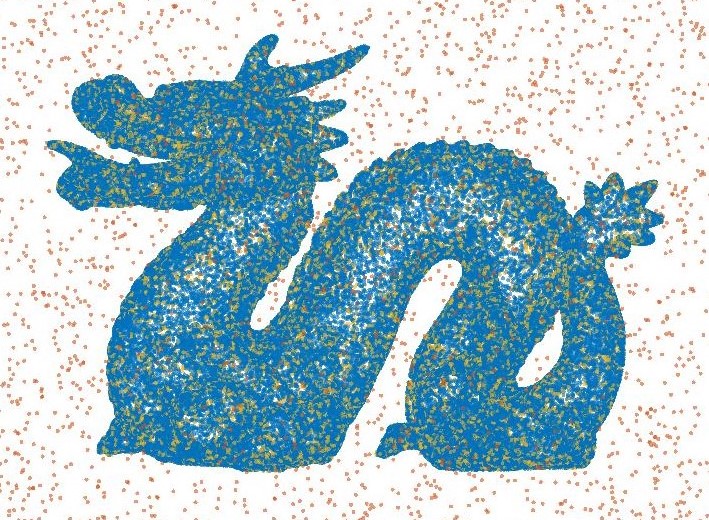}}
\caption{Cleaned Point Cloud as Output of SOR.}
\label{fig:SOR}
\end{figure}

Due to our example with considerable density of noises, SOR is almost invalid. See Fig.~\ref{fig:SOR}.\par
The complexity of SOR approximates to the complexity of  $O(\log^{2}n+f(k)n)$, where $O(\log^{2}n)$ for KT-Tree and both kNN searching on KT-Tree and local statistics will take $O(f(k)n)$, where $f(k)$ denotes function in $k$.

\section{Adaptive Wasserstein Curvature Denoising}
This section contains main original results of this paper. \par
As claimed in the last part of section 2, we believe that the Wasserstein scalar curvature of a symmetric positive-definite matrix $A\in SPD(n)$ implies the quantity of information at $A$. Along the idea of SOR, what the local statistics provides are just positive-definite matrixes as covariances and means. Actually local statistics can be regard as an embedding from the original point cloud into a new point cloud in $SPD(n)$. We can believe that the Wasserstein curvature reflects some relations between points to their neighbors in the new point cloud, which is the start point to distinguish real information and noises. The real information always holds some instinct structures which cause the unevenness of the local points distribution. Hence AWCD is based on a principal hypothesis that the Wasserstein curvatures of the real information are larger than the counterpart of noises. Furthermore, we believe that the Wasserstein curvature of natural noises approximates to which the identify matrix which depicts the an even local distribution. In mathematical words, for a noise point $P_s \in D_0$ embedded onto $\Sigma_s \in SPD(n)$, and a real informational point $P_f$ onto $\Sigma_f$, the Wasserstein scalar curvature  defined as \eqref{Curve scalar} satisfies
\begin{equation}\label{noisecurature }
  \rho(\Sigma_s) \approx \rho(Id(n)) \ll \rho(\Sigma_f).
\end{equation}
\par
\begin{figure}[ht]
\centerline{\includegraphics[width=0.8\linewidth]{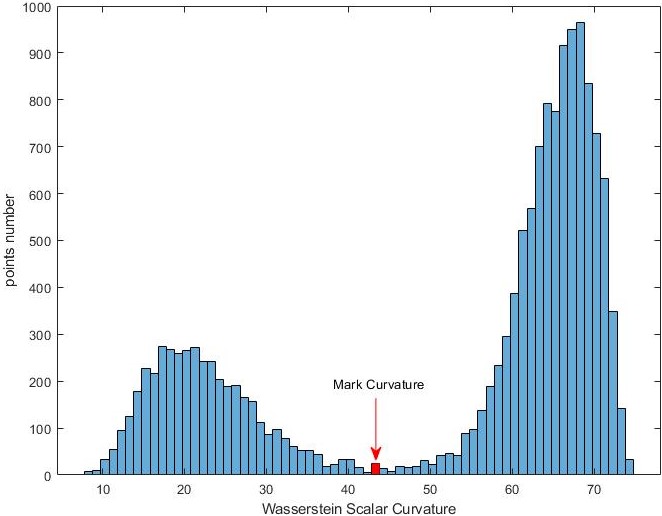}}
\caption{Histogram of Wasserstein Scalar Curvature.}
\label{fig:histogram}
\end{figure}

With such a hypothesis, we design an adaptive algorithm. To achieve adaptivity, we should count the Wasserstein scalar curvature for all points embedded into $SPD(n)$ by a histogram \cite{ref:histogram}. We believe real information and noises will gather around two different 'hills' respectively. Fig.~\ref{fig:histogram} shows the example for the histogram. We can trust that the wave trough in the histogram (The corresponding curvature is called the mark curvature, seeing the red part on Fig.~\ref{fig:histogram}.)  separates the real information against noises. In some cases, if researches tend to choose a mark curvature artificially, what need to do is just omitting this histogram and giving a parameter as the remark curvature.
\par From Fig.~\ref{fig:histogram}, readers may find the Wasserstein curvatures almost remained in a controllable interval, which attributes to \eqref{Scinequal} and our discussions in section 2. \par
Since AWCD involves the Wasserstein scalar curvature which contains merely local information of covariance, the information of bias may be viewed as regular terms sometimes.  In the following, we provide the algorithm of AWCD:
\begin{algorithm}[ht]
\label{alg:AWCD}
\caption{adaptive Wasserstein curvature denoising} 
\hspace*{0.02in} {\bf Input:} 
initial point cloud $D_0$; parameter $k$; \\
\hspace*{0.02in} {\bf Output:} cleaned cloud $D_1$
\begin{algorithmic}[1]
\State search kNN $N_i$ for each point $P_i$
\State compute local mean $\mu_i = \frac{1}{k}\sum_{j=1}^{k}N_{ij}$ and local covariance $\Sigma_i = (N_i)^TN_i$
\State compute Wasserstein curvature $\rho(\Sigma)$  as \eqref{Curve scalar}
\State count curvatures data by histogram and determine the mark curvature $\rho_0$
\If {$\rho(\Sigma_i)\geq\rho_0$ (or with the regular condition $\phi(P_i-\mu_i)$)} {put $P_i$ into $D_1$}
\EndIf
\State \Return $D_1$
\end{algorithmic}
\end{algorithm}
\par
\begin{figure}[ht]
\centerline{\includegraphics[width=0.75\linewidth]{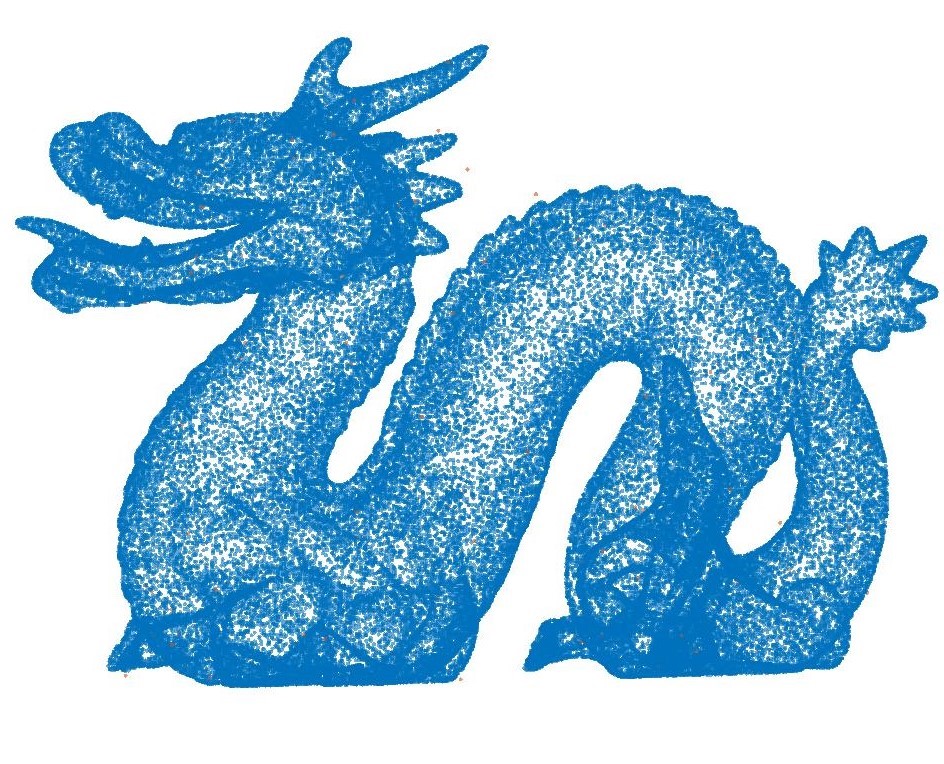}}
\caption{Cleaned Point Cloud as Output of AWCD.}
\label{fig:AWCD}
\end{figure}

 Fig.~\ref{fig:AWCD}. shows the effect of AWCD to process the above example. The efficiency of AWCD is obvious. In this example, AWCD removes almost all noise far from the dragon, and remains almost all real information. The only gap is that a few noises lying on the dragon cause the false positiveness and some true information located on the flat part (such as the abdomen of the dragon) is removed mistakenly. However, such a gap reinforces the dragon structure controversially by moving some inessential points on flat parts while preserving details.  \par
The complexity of AWCD approximates to $O(\log^{2}n+f(k)n+n\log_2n)$, where $\log^{2}n$ for KT-Tree, $f(k)n$ for kNN searching, the curvature computing and final point-wise judgements, meanwhile $O(n\log_2n)$ spent on establishing the histogram by Heapsort \cite{ref:Heapsort}. Therefore, AWCD is still a fast algorithm in common senses. It is remarkable that AWCD is effective for data with dense noises and unsensitive about the unique parameter $k$.

\begin{table*}[ht]
  \centering
  \caption{Denosing effect comparison}
    \begin{tabular}{c|c|ccc|ccc|ccc}
    \hline
    \multicolumn{1}{c|}{\multirow{2}[4]{*}{\tabincell{c}{\textbf{Datasource}\\(Size)}}} & \multicolumn{1}{c|}{\multirow{2}[4]{*}{\tabincell{c}{\textbf{Original}\\ \textbf{SNR}}}} & \multicolumn{3}{c|}{\textbf{ROR}} & \multicolumn{3}{c|}{\textbf{SOR}} & \multicolumn{3}{c}{\textbf{AWCD}} \bigstrut\\
\cline{3-11}      &   & \textbf{TPR} & \textbf{FPR} & \textbf{SNRG} & \textbf{TPR} & \textbf{FPR} & \textbf{SNRG} & \textbf{TPR} & \textbf{FPR} & \textbf{SNRG} \bigstrut\\
    \hline
    \hline
    \multicolumn{1}{c|}{\multirow{2}[2]{*}{\tabincell{c}{\textbf{Stanford Bunny} \\(10000)}}} & \textit{10} & \textit{\textbf{100.00\%}} & \textit{18.10\%} & \textit{452.5\%} & \textit{87.78\%} & \textit{49.00\%} & \textit{79.1\%} & \textit{\textbf{99.28\%}} & \textit{\textbf{4.80\%}} & \textit{\textbf{1968.3\%}} \bigstrut[t]\\
      & \textit{0.1} & \textit{\textbf{100.00\%}} & \textit{100.00\%} & \textit{0.0\%} & \textit{84.44\%} & \textit{\textbf{71.61\%}} & \textit{\textbf{17.9\%}} & \textit{\textbf{99.93\%}} & \textit{90.53\%} & \textit{10.4\%} \bigstrut[b]\\
    \hline
    \multicolumn{1}{c|}{\multirow{2}[2]{*}{\tabincell{c}{\textbf{Duke Dragon} \\(100000)}}} & \textit{100} & \textit{\textbf{100.00\%}} & \textit{9.60\%} & \textit{941.7\%} & \textit{82.29\%} & \textit{51.90\%} & \textit{58.6\%} & \textit{\textbf{100.00\%}} & \textit{\textbf{2.90\%}} & \textit{\textbf{3348.3\%}} \bigstrut[t]\\
      & \textit{1} & \textit{\textbf{100.00\%}} & \textit{100.00\%} & \textit{0.0\%} & \textit{82.12\%} & \textit{70.16\%} & \textit{17.1\%} & \textit{\textbf{99.88\%}} & \textit{\textbf{2.40\%}} & \textit{\textbf{4054.8\%}} \bigstrut[b]\\
    \hline
    \multicolumn{1}{c|}{\multirow{2}[2]{*}{\tabincell{c}{\textbf{Armadillo }\\(50000)}}} & \textit{10} & \textit{\textbf{62.26\%}} & \textit{0.54\%} & \textit{114.3} & \textit{88.06\%} & \textit{63.46\%} & \textit{38.8\%} & \textit{34.84\%} & \textit{\textbf{0.01\%}} & \textit{\textbf{347.38}} \bigstrut[t]\\
      & \textit{1} & \textit{\textbf{62.66\%}} & \textit{0.41\%} & \textit{153.34} & \textit{87.99\%} & \textit{69.62\%} & \textit{26.4\%} & \textit{34.84\%} & \textit{\textbf{0.09\%}} & \textit{\textbf{404.14}} \bigstrut[b]\\
    \hline
    \multicolumn{1}{c|}{\multirow{2}[2]{*}{\tabincell{c}{\textbf{Lucy}\\(50000)}}} & \textit{100} & \textit{\textbf{100.00\%}} & \textit{6.00\%} & \textit{1566.6\%} & \textit{80.71\%} & \textit{51.80\%} & \textit{55.8\%} & \textit{\textbf{99.99\%}} & \textit{\textbf{0.80\%}} & \textit{\textbf{123.99}} \bigstrut[t]\\
      & \textit{10} & \textit{\textbf{100.00\%}} & \textit{35.50\%} & \textit{181.7\%} & \textit{80.65\%} & \textit{66.44\%} & \textit{21.4\%} & \textit{\textbf{99.99\%}} & \textit{\textbf{2.86\%}} & \textit{\textbf{3396.1\%}} \bigstrut[b]\\
    \hline
    \multicolumn{1}{c|}{\multirow{2}[2]{*}{\tabincell{c}{\textbf{Happy Buddha}\\(50000)}}} & \textit{100} & \textit{\textbf{100.00\%}} & \textit{6.80\%} & \textit{1370.6\%} & \textit{77.82\%} & \textit{56.60\%} & \textit{37.5\%} & \textit{\textbf{99.92\%}} & \textit{\textbf{2.40\%}} & \textit{\textbf{4063.3\%}} \bigstrut[t]\\
      & \textit{10} & \textit{\textbf{100.00\%}} & \textit{13.86\%} & \textit{621.5\%} & \textit{78.24\%} & \textit{64.76\%} & \textit{20.8\%} & \textit{\textbf{99.93\%}} & \textit{\textbf{1.88\%}} & \textit{\textbf{5215.6\%}} \bigstrut[b]\\
    \hline
    \end{tabular}
  \label{tab:comparison}
\end{table*}
\section{Digital Experiments}

In this section, we use ROR, SOR and AWCD to denoise polluted data sets with different densities of noises. The point cloud data sets are from the Stanford 3D scanning repository, including Bunny, Dragon, Armadillo, Lucy and Happy Buddha. For each data set, we attach noises of two densities  respectively. The densities of noises are implied as the original signal-noise rate (SNR). In order to show the influences of data size, we down sample the original data sets in distinct scales.  \par
To compare the denoising effects of different approaches, we adopt three criteria of true positive rate (TPR), false positive rate (FPR) and signal-noise rate growing (SNRG). TPR describes the accuracy to preserve the points from the unpolluted set.  FPR describes the success rate to remove noises. To evaluate the denoising effects, SNRG is the most direct criteria, which explicates the promotion of SNR through processing. For any polluted point cloud  $D_0 = D\cup N$, where $D$ is the points set of real information and $N$ is the set of noises, we attain the cleaned cloud $D_1$ by the denoising algorithm, then
\begin{align}
  {\rm TPR} & = \frac{|D_1\cap D|}{|D|} ,\\
  {\rm FPR} & = 1-\frac{|D_1\cap N|}{|N|}, \\
  {\rm SNRG} & = \frac{|D_1\cap D|}{|D_1\cap N|}\cdot\frac{|N|}{|D|}-1,
 \end{align}
 where $|\cdot|$ denotes the
cardinality or size of a finite set.
Intuitively, higher TPR, SNRG and lower FPR mean the better ability of an algorithm to distinguish real information against noise.
Therefore in Tab.~\ref{tab:comparison}, for each comparison experiment, we highlight the lowest FPR, the highest TPR and the SNRG over 99\%.\par
Tab.~\ref{tab:comparison} explicitly shows the advantage of AWCD, compared with ROR and SOR. In general, AWCD can remove almost all noises and meanwhile preserves the real information better, except for Armadillo.

\section{Conclusion}
To sum up, this paper introduces AWCD, a new efficient denoising algorithm for point cloud. By involving the Wasserstein curvatures, this method can denoise point cloud with dense noise adaptively. We list main advantages and disadvantages of this algorithm:
\begin{itemize}
\item Adaptivity,
\item Fast,
\item Efficient to noises in high density,
\item Lack of the explainablity.
\end{itemize}
\par Although we have no idea about why the Wasserstein curvature has such benefits for point cloud denoising, AWCD algorithm may be regarded as a positive try to involve the curvature information into data processing. AWCD implies the importance of the local statistics and Wasserstein distance.  An advancer edition of AWCD and its explainablity requirs further study.

\section*{Acknowledgements}
This research was funded by National Key Research and Development Plan of China, No.
2020YFC2006201.
\bibliographystyle{IEEEtran}
\bibliography{references.bib}

\end{document}